\documentclass{article}

\usepackage{microtype}
\usepackage{graphicx}
\usepackage{subcaption}
\usepackage{booktabs} 

\usepackage{hyperref}
\usepackage{balance}

\usepackage[preprint]{icml2026}

\usepackage{amsmath}
\usepackage{amssymb}
\usepackage{mathtools}
\usepackage{amsthm}

\usepackage[capitalize,noabbrev]{cleveref}

\theoremstyle{plain}

\theoremstyle{definition}

\theoremstyle{remark}

\usepackage[textsize=tiny]{todonotes}

\icmltitlerunning{Submission and Formatting Instructions for ICML 2026}

\begin{document}

\twocolumn[
  \icmltitle{OUI as a Structural Observable: Towards an \\ Activation-Centric View of Neural Network Training}



  \icmlsetsymbol{equal}{*}

\begin{icmlauthorlist}
  \icmlauthor{Alberto Fernández-Hernández}{upv}
  \icmlauthor{Jose I. Mestre}{upv}
  \icmlauthor{Cristian Pérez-Corral}{upv}
  \icmlauthor{Manuel F. Dolz}{uji}
  \icmlauthor{Jose Duato}{openchip}
  \icmlauthor{Enrique S. Quintana-Ortí}{upv}
\end{icmlauthorlist}

\icmlaffiliation{upv}{Universitat Politècnica de València, Valencia, Spain}
\icmlaffiliation{uji}{Universitat Jaume I, Castelló de la Plana, Spain}
\icmlaffiliation{openchip}{Openchip \& Software Technologies S.L., Spain}

\icmlcorrespondingauthor{Alberto Fernández-Hernández}{a.fernandez@upv.es}

  \icmlkeywords{Machine Learning, ICML}

  \vskip 0.3in
]



\printAffiliationsAndNotice{}  

\begin{abstract}
Activation functions are what make deep networks expressive: without them, the model collapses to a linear map. Yet we still evaluate training mostly from the outside, through loss, accuracy, return, or final calibration, while the internal structural evolution of the network remains largely unobserved. In this paper, we argue that the Overfitting--Underfitting Indicator (OUI) should be understood as a first practical observable of that internal structure. Across our recent results, OUI consistently appears as an early, label-free, activation-based signal that reveals whether a network is entering a poor or promising training regime before convergence. In supervised learning, it anticipates weight decay regimes; in reinforcement learning, it discriminates learning-rate regimes early in PPO actor--critic; and in online control, it can drive layer-wise weight decay adaptation. Read together with recent evidence that activation patterns tend to stabilize earlier than parameters, these results suggest a broader research direction: an activation-centric theory of training dynamics. OUI is becoming an empirical foothold toward this theory.
\end{abstract}

\section{Introduction}

Neural Networks (NNs) owe their expressive power to their activations. Once nonlinearity disappears, so does the ability to partition the input space into distinct functional regions. For models that use ReLU activations or related functions such as GELU or SiLU, this partition is not merely metaphorical but corresponds to a well-defined geometric structure: activation patterns determine which local affine map is applied to each input \cite{hartmann_studying_2021, perez-corral_regime_2026}. If we want to understand how NNs learn, then understanding how these activation structures emerge, stabilize, and fail is central.

The problem is that our standard training diagnostics rarely look there. Representation-similarity methods such as CKA \cite{kornblith_similarity_2019} compare activations across checkpoints or layers but are not designed as online training signals. Dead neuron tracking monitors collapse but not the richer structure of how living units discriminate among inputs. Loss and accuracy arrive late and compress away most of the internal story. They tell us whether training worked, not what structural regime the model entered on the way there. This is precisely the gap that makes hyperparameter search expensive, late failure common, and training dynamics hard to interpret.

Our starting point is the Overfitting--Underfitting Indicator (OUI), an activation-based, label-free metric originally introduced as a practical early signal for weight decay selection \cite{fernandez-hernandez_oui_2025}. Since then, OUI has been reformulated in a cheaper batch-based form and shown to separate learning-rate regimes early in PPO actor--critic \cite{fernandez-hernandez_when_2026}.
In parallel, recent work on activation-pattern dynamics suggests that activation regimes often stabilize substantially earlier than parameters, supporting a two-timescale view of training \cite{perez-corral_regime_2026}. Taken together, these results point to a stronger claim than the one OUI was born with.

This paper makes that claim explicit by arguing that OUI should be understood as a structural observable of learning dynamics. Across results in supervised regularization, reinforcement learning, online control, and activation-pattern convergence, we extract a common activation-centric hypothesis: activation-based observables can reveal an early, actionable, and potentially theory-relevant dimension of training that standard external metrics do not capture directly. Our contribution is therefore a compact synthesis and reinterpretation of existing evidence, together with the practical implications and the main open questions that must be addressed if activation-based observables are to become useful not only for practice, but also for theory and benchmarking.

What is new in this paper is therefore the research hypothesis that emerges when these results are read together. Across supervised learning, reinforcement learning, online regularization, and activation-pattern convergence studies, the recurring signal is the same: internal activation structure becomes informative early, before conventional external metrics have fully revealed the fate of the run. Our contribution is to make that common claim explicit, to identify it as the basis of an activation-centric view of training, and to formulate the next questions needed to turn that view into a genuine research program.

\section{What OUI Has Already Revealed}

Before discussing what OUI has shown across settings, it is useful to state what OUI actually measures. We adopt the batch-based formulation introduced in \cite{fernandez-hernandez_when_2026}, since it is compact, cheap to evaluate online, and well suited to monitoring training dynamics. Consider a layer $l$ with $d_l$ scalar activation units, and let $\mathcal{B}_t=\{x_1,\dots,x_B\}$ be the training batch at training step $t$. For each sample $x_b$ and neuron $n$, let $a^{(l)}_n(x_b;\theta_t)$ denote the corresponding preactivation under parameters $\theta_t$. We define the binary activation mask $m^{(l)}_{b,n}(t)=\mathbf{1}\!\left\{a^{(l)}_n(x_b;\theta_t)>0\right\},$ which simply records whether neuron $n$ is active for sample $x_b$. For each neuron, we then count how many samples in the batch activate it through $s^{(l)}_n(t)=\sum_{b=1}^{B} m^{(l)}_{b,n}(t).$ From this, we compute the minority count $u^{(l)}_n(t)=\min\!\left(s^{(l)}_n(t),\, B-s^{(l)}_n(t)\right),$ which is large when the neuron splits the batch in a balanced way and small when it is almost always on or almost always off. The batch-based OUI of module $l$ at step $t$ is then
\begin{equation}
\mathrm{OUI}_i(t)=
\frac{1}{d_l}
\sum_{j=1}^{d_l}
\frac{u^{(l)}_j(t)}{\lfloor B/2 \rfloor}.
\end{equation}

\begin{figure}[h]
    \centering
    \includegraphics[width=\linewidth]{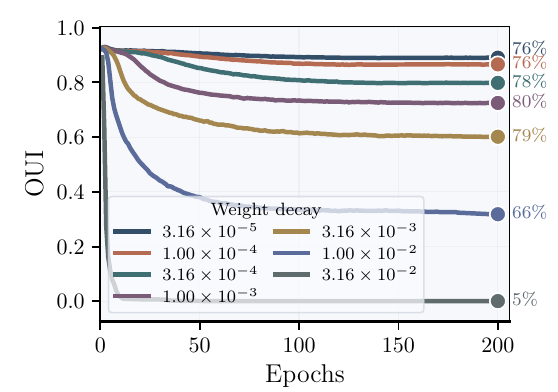}
\caption{OUI trajectories for a DenseNet-BC-100 on CIFAR-100 under seven logarithmically spaced weight-decay values. OUI quickly enters stable regimes, and by $\sim$15\% of training the trajectories are already separated, anticipating which settings will generalize well or lead to under- or overfitting (validation accuracy in \% shown on the right).}
\label{fig:oui-wd}
\end{figure}

Operationally, OUI is high when many units partition the probe batch in a roughly balanced manner, and low when the module becomes biased, with many units behaving almost uniformly across inputs. In this sense, OUI looks directly at how the internal activation structure is being used. That is the key shift in perspective throughout this paper: OUI is treated as an observable of a hidden structural dimension of learning.

The original OUI result was practical but already suggestive \cite{fernandez-hernandez_oui_2025}. OUI was first introduced as a scalar quantity derived from activation patterns and shown to anticipate whether a weight decay choice would push a model toward underfitting, overfitting, or a better regularization balance (see Figure \ref{fig:oui-wd}). The key fact was not simply that OUI correlated with final performance. It was that it became informative early, often before standard external metrics had settled. That result already hinted that activation structure contains useful information about the future trajectory of training.

\begin{figure}[!h]
    \centering
    \includegraphics[width=\linewidth]{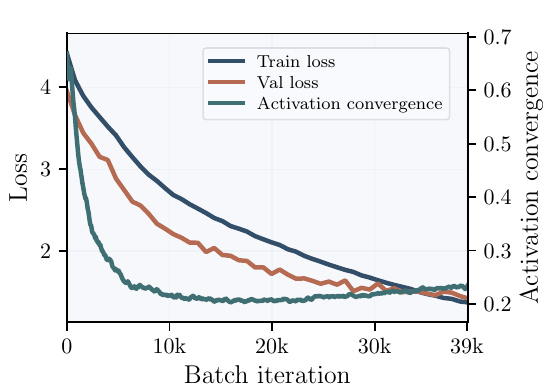}
\caption{Training and validation loss together with an activation-pattern convergence metric for a ViT-B16 on CIFAR-100. The activation metric saturates much earlier, while losses continue improving, illustrating that structural organization stabilizes before parameter refinement.}
\label{fig:regime-change}
\end{figure}

\begin{figure}[!b]
    \includegraphics[width=0.95\linewidth]{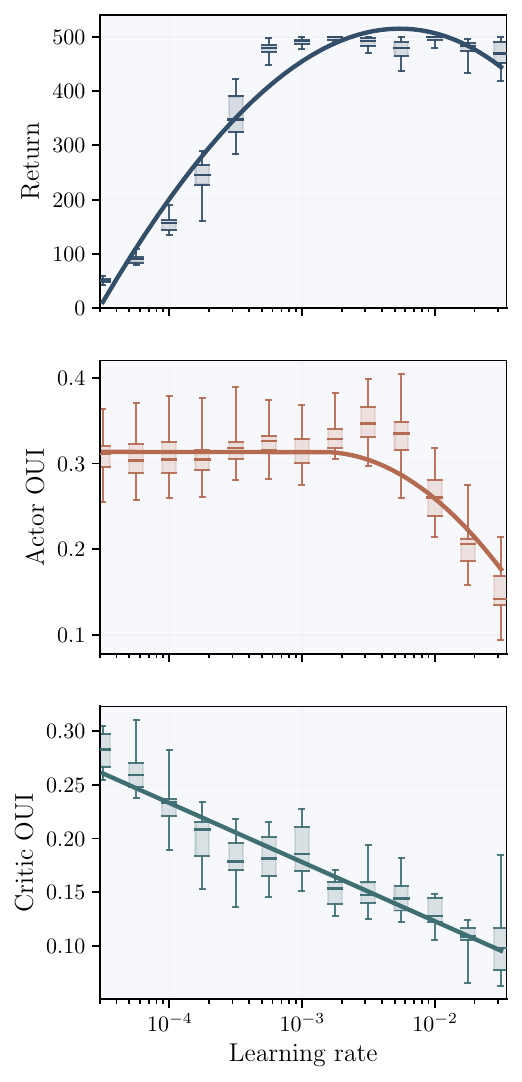}
\caption{Return, actor OUI, and critic OUI across learning rates for PPO on CartPole-v1. While performance shows a peaked dependence on the learning rate, actor and critic exhibit distinct OUI profiles, highlighting role-dependent activation structure and enabling early identification of good regimes.}
\label{fig:oui-ppo}
\end{figure}

The second step mattered even more. In PPO actor--critic, OUI was reformulated in the batch-based form above, making it cheaper and more stable to evaluate online \cite{fernandez-hernandez_when_2026}. In that setting, OUI measured at only a small fraction of training was sufficient to discriminate between learning-rate regimes, and it did so competitively against signals practitioners normally monitor, including early return and PPO-specific quantities. Just as important, the actor and critic did not share the same preferred structural regime: good critics occupied an intermediate OUI band, while strong actors tended to run at higher OUI values (see Figure \ref{fig:oui-ppo}). This is a useful correction. It suggests that OUI is pointing to a structural property whose good range depends on architectural role and training context.

Finally, the regime-change \cite{perez-corral_regime_2026} line gives these observations a broader conceptual frame. Building on the geometry of piecewise-linear networks, that work argues and empirically supports a two-timescale interpretation of training: activation-pattern changes tend to decay earlier than parameter updates, so a neural network may spend a significant late stage refining weights within comparatively stable activation regimes (see Figure \ref{fig:regime-change}). OUI fits naturally inside that picture. If activation structure settles early enough to matter, then an activation-based observable should be able to detect meaningful shifts in training before convergence. OUI is, at the very least, one concrete candidate for doing so.

\section{What This Suggests in Practice}

The most immediate consequence of this view is early screening: if OUI separates promising and poor regimes before convergence, then it could serve as an internal early-warning signal in foundation-model pretraining, where late-stage failures are catastrophically expensive and where benchmarked training dynamics matter as much as final performance. But stopping there would miss the real point. The deeper consequence is that training can be monitored through an internal structural signal, not only through delayed external outcomes. A good early signal changes what we think we are measuring. Instead of asking only whether a run eventually scores well, we can ask what kind of structural organization is forming inside the network while learning is still underway.

Another implication is functional control. Gradient-based and weight-based adaptation methods act through optimization quantities. OUI acts through activation structure. That difference matters. It means we can regulate training from a signal that is closer to what the network is functionally doing to data, rather than from what the optimizer is numerically doing to parameters \cite{fernandez-hernandez_oui_2025}. Even if OUI is not the final answer, this is already a meaningful shift in perspective.

These considerations also extend to benchmarking. Current benchmarks mostly compare endpoints, but an activation-centric view suggests benchmarking training dynamics directly: when internal trajectories become separable, how stable that separation is across seeds and probe batches, how well early structural signals correlate with final performance, and whether they transfer across architectures, tasks, and hyperparameters. In that sense, OUI is a candidate instrument for training-dynamics benchmarks, particularly in foundation models, where efficient and stable emergence of useful internal regimes matters alongside final performance.

The practical lesson is therefore simple. OUI behaves like an early structural observable. It is cheap enough to monitor, rich enough to be useful, and already informative across settings that differ in objective, architecture, and hyperparameter. This is precisely where our contribution departs from a retrospective summary. We are not merely observing that OUI happened to work in several settings. We are arguing that these settings reveal the same underlying pattern from different angles: activation-based observables can become useful before convergence because training has an internal structural organization that is not captured by endpoint metrics alone. Naming that pattern, and isolating it as a scientific object, is part of the contribution of this paper.

\section{Toward an Activation-Centric Theory of Training}

The broader motivation behind this paper is simple. Activations are not an incidental detail of NNs; they are the mechanism through which expressive nonlinear behavior becomes possible. If that is true, then understanding how activation structure emerges and stabilizes is not just a practical matter of diagnostics. It is part of understanding what the network is learning internally. OUI matters to us not only because it is useful early, but because it suggests that this internal structural layer is measurable, actionable, and perhaps eventually explainable.

If the preceding results are read narrowly, OUI is a helpful engineering heuristic. Read together, however, they support a stronger interpretation: activation structure is not just an internal byproduct of training, but a legitimate layer of description of how learning unfolds. Notably, the underlying pattern is not confined to a single task family or model class, but appears across supervised and reinforcement learning settings as well as CNN, transformer, and MLP-like architectures. This is the central claim that motivates the activation-centric perspective defended here.

The first question is phase structure. If activation regimes stabilize earlier than parameters in many settings \cite{perez-corral_regime_2026}, then training may not be a single homogeneous process. It may have an early structural phase, where regions and gates are still being organized, followed by a later refinement phase, where parameters continue to move inside a more stable structural scaffold. OUI may be useful precisely because it is sensitive to that first phase.

The second question is adequacy. What makes an activation structure good? The answer is unlikely to be ``high OUI'' or ``intermediate OUI'' in the abstract. The PPO asymmetry already warns against that simplification \cite{fernandez-hernandez_when_2026}. What matters is whether the induced activation regime is appropriate for the role of the module, the task, and the stage of training. In foundation models, this may also bear on capability differences, for instance between regimes associated with memorization and those associated with in-context learning or compositional generalization. A theory of activations will have to explain this contextual adequacy.

The third question is theory. ReLU networks are piecewise affine; activation patterns determine local regions and therefore local functional behavior. Prior work \cite{hartmann_studying_2021} has shown that these patterns evolve with architecture-dependent regularity and that they encode meaningful changes in expressivity during training. OUI, in turn, compresses part of that internal behavior into a scalar that is early and actionable. The next step is to connect such observables to more formal notions of stability, regime transitions, and eventually generalization. This connection remains unresolved, but with metrics like OUI, now appears to be within reach.

The final question is design. If activation structure is a first-class object, then training procedures can be redesigned around it. This includes early hyperparameter pruning, activation-driven regularization, two-phase optimization once regimes stabilize, and new dynamic benchmarks that evaluate not only where models end, but how they get there. In that broader picture, OUI is best understood as a beginning: one of the first compact signals that makes such a theory empirically approachable.

\section{Conclusion}
OUI reveals that activation structure can be observed early enough to matter. That observation changes the role of activations from hidden byproduct to explicit object of study. If that shift holds, then an activation-centric theory of training is not a speculative luxury. The purpose of this paper has been to make that next step visible.

\section*{Impact Statement}

Activation-based observables such as OUI may improve training efficiency by identifying poor regimes early, but they also raise concrete risks. First, OUI is most naturally defined for ReLU-like activation patterns and may become less informative, or require reformulation, under other activations. Second, if used as an online control signal, it can induce a feedback-loop confound, where training adapts to manipulate the observable itself rather than the underlying learning dynamics. These limitations make careful validation essential, particularly in large-scale foundation-model settings.

\section*{Acknowledgements}
This research was funded by the projects PID2023-146569NB-C21 and PID2023-146569NB-C22 supported by MICIU/AEI/10.13039/501100011033 and ERDF/UE. Alberto Fernández-Hernández was supported by the predoctoral grant PREP2023-001826 supported by MICIU/AEI/10.13039/501100011033 and ESF+. Jose I. Mestre was supported by the predoctoral grant ACIF/2021/281 of the \emph{Generalitat Valenciana}. Cristian Pérez-Corral received support from the \textit{Conselleria de Educación, Cultura, Universidades y Empleo} (reference CIACIF/2024/412) through the European Social Fund Plus 2021–2027 (FSE+) program of the \textit{Comunitat Valenciana}. Manuel F. Dolz was supported by grant {\small CNS2025-165098} funded by {\small MICIU/AEI/10.13039/501100011033} and by the Plan Gen--T grant {\small CIDEXG/2022/013} of the \emph{Generalitat Valenciana}.


\balance

\bibliography{references}
\bibliographystyle{icml2026}

\end{document}